\documentclass{article}

\usepackage[dblblindworkshop,final]{neurips_2026}
\providecommand{\workshoptitle}[1]{}
\workshoptitle{Agents in the Wild: Safety, Security, and Beyond}

\usepackage[utf8]{inputenc}
\usepackage[T1]{fontenc}
\usepackage{hyperref}
\usepackage{url}
\usepackage{booktabs}
\usepackage{amsmath}
\usepackage{amssymb}
\usepackage{amsfonts}
\usepackage{microtype}
\usepackage{enumitem}
\usepackage{textcomp}
\usepackage{algorithm}
\usepackage{algpseudocode}
\usepackage{cleveref}
\usepackage{tikz}
\usetikzlibrary{arrows.meta,positioning,fit,backgrounds,shapes.misc,calc}
\usepackage[svgnames,x11names,dvipsnames]{xcolor}

\definecolor{darkblue}{rgb}{0, 0, 0.5}
\hypersetup{colorlinks=true, citecolor=darkblue, linkcolor=darkblue, urlcolor=darkblue}

\title{AdMem: Expectation-Grounded Procedural Memory for Long-Horizon Task-Solving Agents}

\author{
  Runzhe Wang \\ Princeton University \And
  Huilin Lu \\ Amazon \And
  Shengjie Liu \\ Amazon \And
  Li Dong \\ Amazon \And
  Zhihuai Zhu \\ Arm
}

\begin{document}

\maketitle

\begin{abstract}
An agent that carries memory across tasks is only as good as the entries it keeps. We begin from an observation that is easy to reproduce and uncomfortable for the field: naively accumulating cross-task experience can make an agent \emph{worse} than one with no memory at all. On AgentBoard, a workflow-induction memory solves fewer tasks than a memoryless ReAct agent in seven of eight domains, and a per-step procedural store with no surrounding structure loses half of ReAct's completions on Jericho. We argue that the missing ingredients are per-action credit assignment and an explicit life cycle for stored entries, and present AdMem, which supplies both. Before each action the actor commits to an expected outcome; a concurrently running critic scores the realized outcome against that expectation and writes a procedural entry carrying a binary reward and a reflection, so credit lands on the action that earned it and no environment reward is required. Each entry then carries a scalar utility, updated online under a noisy-OR reward model, that ranks retrieval jointly with context similarity and drives eviction and merging, giving the store a forgetting and consolidation policy instead of unbounded growth. Across 707 tasks in eight AgentBoard domains AdMem solves 441 against 388 for ReAct and 312 for the workflow-induction baseline --- a pooled gain significant under a two-sided Fisher exact test, concentrated in the domains that supply transferable cross-task structure. All memory stays human-readable text, which is what makes the failures above diagnosable --- and, as we discuss, also makes the store an auditable but attackable surface.
\end{abstract}

\section{Introduction}
Large Language Models (LLMs) \citep{brown2020languagemodelsfewshotlearners, JMLR:v24:22-1144, deepseekai2025deepseekr1incentivizingreasoningcapability, touvron2023llama2openfoundation} have improved rapidly at reasoning and tool use, yet still struggle with long multi-turn tasks that require remembering, organizing, and applying knowledge across sessions or large inputs \citep{zhang2024surveymemorymechanismlarge}. Two research directions have emerged \citep{wu2025humanmemoryaimemory}: \emph{architectural memory}, built into the network through added layers or modules \citep{xu2023parameterefficientfinetuningmethodspretrained}, and \emph{symbolic or textual memory} that the model reads and writes explicitly. This paper is about the second, and specifically about a question the second direction makes it possible to ask: not how much experience an agent can store, but which of it is worth keeping.

Parametric memory extends the model's effective context: End-to-End Memory Networks \citep{sukhbaatar2015endtoendmemorynetworks} introduced differentiable attention over an external memory, and later work \citep{bergesmemory, behrouz2024titanslearningmemorizetest, munkhdalai2024leavenocontextbehind} learns long-term memory modules that retain historical context while keeping training parallelizable. These buy speed and differentiability at the cost of inspectability: what has been remembered is distributed over weights or a compressive state, so it cannot be read, attributed, or deleted item by item. That trade-off is decisive here, because the central failure we report is an agent storing the wrong experience and re-using it: diagnosing it required reading the offending entries, and fixing it required a policy that can remove a specific one. We therefore keep every entry as readable text.

Textual stores are inspectable and tool-friendly, but need well-designed policies to be effective and bounded. Prior systems fine-tune models to read and write knowledge triples \citep{modarressi2025memllmfinetuningllmsuse, modarressi2024retllmgeneralreadwritememory}, foreground human control over adding, merging, and deleting entries \citep{huang2023memorysandboxtransparentinteractive}, page virtual context with explicit CRUD \citep{packer2024memgptllmsoperatingsystems}, consolidate units extracted from running conversations \citep{chhikara2025mem0buildingproductionreadyai, zhou2025mem1learningsynergizememory}, decay episodic histories on a forgetting curve \citep{zhong2023memorybankenhancinglargelanguage}, and compress context by learned overwriting or sub-task chunks \citep{yu2025memagentreshapinglongcontextllm, hu2025hiagent}. As a group they already hold most of the vocabulary a memory life cycle needs --- writing, consolidation, decay, deletion --- but instantiate it for facts about the world and the user, where an entry's value is roughly its correctness. Our entries are recommendations about \emph{how to act}, whose value is whether following them helped, and that is not observable at write time.

That gap is the domain of \emph{procedural} memory: in the classical cognitive division \citep{tulving1985howmany, cohen1997memory, wu2025humanmemoryaimemory} it holds knowledge of how to do things rather than the semantic and episodic content above. Cast into the language-agent setting \citep{sumers2023cognitive, gao2025survey}, the LLM is the decision maker and the prompt is its control flow, so procedural memory is whatever text conditions the next action. Agent Workflow Memory (AWM) \citep{wang2024agentworkflowmemory} induces reusable workflows from past successes in web navigation \citep{deng2023mind2webgeneralistagentweb, zhou2024webarenarealisticwebenvironment}; Mem$^p$ \citep{fang2025mempexploringagentprocedural} distills experience into granular instructions and higher-level templates \citep{xie2024travelplannerbenchmarkrealworldplanning, shridhar2021alfworldaligningtextembodied}; Buffer of Thoughts \citep{yang2024bufferthoughtsthoughtaugmentedreasoning} stores distilled thought templates; and MIRIX \citep{wang2025mirixmultiagentmemoryllmbased} and Agent KB \citep{tang2025agent} organize structured memory types or strategy--log pairs under a multi-agent controller \citep{mialon2023gaiabenchmarkgeneralai, jimenez2024swebenchlanguagemodelsresolve}. A parallel line writes memory from evaluated outcomes rather than successes alone: Reflexion \citep{shinn2023reflexionlanguageagentsverbal} turns task failure into a verbal reflection, ExpeL \citep{zhao2024expel} accumulates insights from successful and failed trajectories, Generative Agents \citep{park2023generativeagentsinteractivesimulacra} maintains a reflection hierarchy, and the most recent work --- ReasoningBank \citep{ouyang2025reasoningbank}, Memento \citep{zhou2025memento}, and agentic context engineering \citep{zhang2025ace} --- makes reusable reasoning strategy the stored unit.

AdMem shares its goal with that line and differs in two specific ways that the rest of the paper is about. \textbf{Where credit is assigned.} These systems evaluate a trajectory or a task and write an entry from the verdict, so the reward is task-level and arrives once, after many actions --- the credit-assignment problem in its usual form. AdMem instead requires the actor to state, before each action, what it expects that action to accomplish, and has a critic score the realized outcome against that expectation as soon as the relevant observations arrive. Credit lands on the action that earned it, with no environment reward needed. \textbf{What happens to an entry afterwards.} Where these systems do keep post-write bookkeeping --- helpfulness counters, deduplication, a value per stored case --- it is updated from the task-level verdict and is a separate quantity from the score retrieval ranks by. AdMem gives each procedural entry one scalar utility with an explicit probabilistic meaning --- the probability that following it makes the current action succeed --- estimated online from the per-action rewards of the paragraph above, and ranks retrieval by that same quantity, with eviction and merging following from the retrieval statistics it drives. The store thus has a life cycle --- write, evaluate, consolidate, forget --- closed by one estimated number rather than a write path plus side counters.

This matters more than it may appear, because unmanaged accumulation is not merely suboptimal. In our experiments a workflow-induction memory ends up \emph{below} a memoryless ReAct control in seven of eight AgentBoard domains, and an ablation of our own system that keeps the procedural store but removes the planning and scoping structure around it loses half of ReAct's completions on Jericho. An independent streaming benchmark reports the same for workflow-induction memory in a four-backbone evaluation \citep{evomemory2025}. A memory that cannot be evaluated and cannot be forgotten is a liability whenever the domain supplies less transferable structure than the write policy assumes.

Our contributions are:
\begin{itemize}[itemsep=0pt, topsep=2pt, parsep=0pt, partopsep=0pt]
   \item \textbf{Expectation-grounded credit assignment} for procedural memory: the actor commits to an expected outcome before acting, and a concurrent critic turns the expectation--outcome comparison into a per-action binary reward plus a reflection, with no environment reward required (\Cref{sec:pipeline}).
   \item \textbf{An explicit life cycle for the long-term store}: a per-entry utility updated online under a noisy-OR reward model, ranking retrieval jointly with context similarity and driving frequency-based eviction and agreement-based merging, so the store forgets and consolidates rather than growing without bound (\Cref{sec:memory-management}).
   \item \textbf{A short-term memory scoped as a sub-goal stack}, cleared at task boundaries, bounding the actor's context and giving the critic well-delimited units to evaluate (\Cref{sec:stm}).
   \item \textbf{An empirical result with a negative half}: across 707 tasks in eight domains, cross-task procedural memory without these mechanisms is significantly \emph{worse} than no memory at all, while AdMem is significantly better, with gains concentrated where cross-task structure exists (\Cref{sec:experiments}).
\end{itemize}

\section{Methodology}
\subsection{Setting, notation, and framework}
An agent solves a series of tasks by interacting with an environment in turns; in round $t$ it emits an action $a_t$ and the environment returns an observation $o_t$. The environment may comprise human users, other agents, or tool-calling infrastructure. Each task is its own environment episode, but the agent's memory is \emph{not} reset between tasks --- the setting of a deployed assistant, and the reason a single lifelong horizon, rather than a task, is the right unit of analysis. The agent carries a memory $s_t$ and acts by conditioning an LLM on a context $c_t$ assembled from that memory, $a_t\sim\pi_{\text{LLM}}(\cdot\mid c_t)$. Vanilla baselines for multi-turn benchmarks \citep{ma2024agentboard} take $c_t=s_t=[a_1,o_1,\dots,a_t,o_t]$, the full trajectory, which stresses long-context reasoning as tasks grow. Prior work responds along two axes: \emph{context compaction}, keeping $s_t$ short enough for one window, and \emph{retrieval}, placing only part of the memory into $c_t$. These correspond to short-term and long-term memory, and AdMem uses both: $s_t = (s_t^{\text{short}}, M)$, where $s_t^{\text{short}}$ is per-task and $M$ persists across tasks.

\begin{figure}[t]
\centering
\resizebox{0.80\linewidth}{!}{
\begin{tikzpicture}[
  node distance=10mm and 20mm,
  title/.style   ={font=\bfseries\large},
  label/.style   ={font=\bfseries\small},
  small/.style   ={font=\small},
  tinytext/.style={font=\scriptsize},
  box/.style={draw, rounded corners, thick, minimum width=30mm, minimum height=10mm, align=center},
  >=Latex,
  envbox/.style   ={box, fill=gray!12},
  actorbox/.style ={box, fill=blue!10},
  criticbox/.style={box, fill=orange!12},
  membox/.style   ={box, fill=green!10},
  heap/.style     ={draw, rounded corners=1.5pt, thick, fill=BlueViolet!10},
  stack/.style    ={draw, rounded corners=1.5pt, thick,fill=BlueViolet!10},
  outer/.style    ={draw, rounded corners=3pt, thick, fill=black!2}
]

% Nodes
\node[envbox,font=\bfseries, minimum width=2cm] (user)    {Environment};

\draw[outer] (3.2,0.8) rectangle (13.7,-5);
\node[actorbox,anchor=north, minimum width=4cm,
  minimum height=5.2cm, align=center,
  anchor=north] at (5.5,0.5) (actor) {};
\node[title] at (5.5,0) {Actor};
\node[criticbox,anchor=north, font=\bfseries, minimum width=2cm] at (11.5,-0.3) (critic) {Critic};
\node[membox,anchor=north, font=\bfseries, minimum width=2cm] at (11.5,-2.8) (memory) {Long-Term\\Memory Store};

\node[anchor=north, small] at (5.5, -0.7) {\shortstack[l]{\textless Task instructions\textgreater\\\textless Task tools\textgreater\\\textless Memory tools\textgreater\\\textless Task STM\textgreater \\\textless Retrieved LTM\textgreater}};

\draw[-] (4, -0.5) to (7,-0.5);
\draw[-] (4, -2.7) to (7,-2.7);
\node[heap, label] at (5.5, -3.2) {Semantic Heap};
\node[stack, label] at (5.5, -4.0) {Episodic Stack};

%coordinate
%\draw[step=1cm, gray!40] (-1,-5) grid (14,1);
%\draw[thick,->] (-1,0) -- (14.2,0) node[right] {$x$};
%\draw[thick,->] (0,-5) -- (0,1.2) node[above] {$y$};

\draw[->, thick, bend left=3] ($(user.east) + (0.1,0.1)$) to node[midway, above]{Observation} ($(3.2,0) + (-0.1,0.1)$); 
\draw[<-, thick, bend right=3] ($(user.east) + (0.1,-0.1)$) to node[midway, below]{Action} ($(3.2,0) + (-0.1,-0.1)$);

\draw[->, thick] ($(7.5, -1)$) to node[midway, above, font=\small]{\shortstack[l]{Task STM\\Action\\Expectation\\Observation}} ($(critic.west)+(0, -0.2)$);  
\draw[->, thick] (critic) to node[midway, right, font=\small]{\shortstack[l]{Procedural entries\\+ per-action rewards}} (memory);
\draw[->, thick, bend left=10] ($(7.5, -2.9)$) to node[midway, above = 0mm]{\shortstack{Semantics\\\& episodics}} (memory);  
\draw[<-, thick, bend right=10] ($(7.5, -3.7)$) to node[midway, below]{\shortstack{Retrieved\\LTM}} (memory);      

\end{tikzpicture}}
\caption{Information flow between the environment, the actor, the critic, and the long-term memory store. The three modules run in separate threads: the actor never blocks on the critic, so memory writing and evaluation overlap with acting. The actor's short-term memory (STM) holds the task instructions, the task and memory tool specifications, and the current task's short-term trace, with environment facts surfaced through a priority-ranked \emph{semantic heap} and event summaries through a scope-ordered \emph{episodic stack} (\Cref{sec:stm}); retrieved long-term memory (LTM) enters the same context. Every action the actor sends to the critic carries the expectation the actor committed to before acting, which is what the critic scores the realized outcome against; the resulting per-action reward travels to the store alongside the new procedural entry and updates the utilities of the entries that were retrieved for that action (\Cref{sec:memory-management}).}
\label{fig:interaction_diagram}
\end{figure}
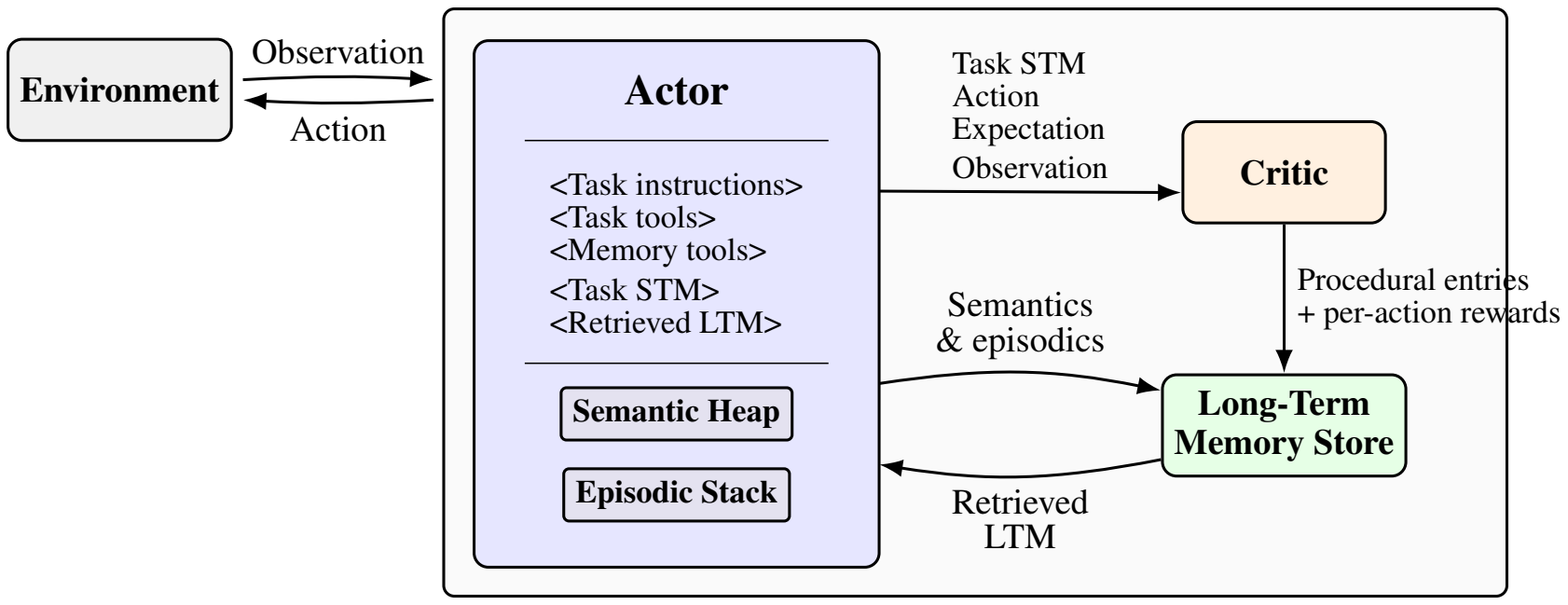
AdMem runs three modules concurrently (\Cref{fig:interaction_diagram}), only the first of which is a decision-making agent. The \textbf{actor} is the LLM agent that interacts with the environment; it maintains $s_t^{\text{short}}$, updated every turn and cleared when the task ends, by which point anything worth keeping has been written to $M$. The \textbf{long-term memory store} is a retrieval index over $M$ that also owns the management policy of \Cref{sec:memory-management}; it performs no reasoning of its own beyond LLM-mediated updating of semantic entries. The \textbf{critic} is a second LLM instance on the memory write path: it caches each action together with the actor's stated expectation, waits until the relevant observations have arrived, then scores the action and writes the resulting procedural entry. It never emits an environment action.

The store holds three kinds of entry, $M = M_S \oplus M_E \oplus M_P$, following the standard cognitive division \citep{tulving1985howmany, cohen1997memory}: $M_S$, \textbf{semantic} --- facts and general knowledge about the environment and the tasks, such as what a tool returns or what a field name means; $M_E$, \textbf{episodic} --- summaries of past events in the agent's own interaction history; and $M_P$, \textbf{procedural} --- decision-making guidance extracted from past actions, recording for each action taken and outcome observed whether the outcome met the stated expectation and advanced the task, and what a better action would have been. Semantic and episodic entries are comparatively easy to form: an LLM summarizes facts and events as the task runs. Procedural entries are harder for two reasons. First, an action cannot be judged until the observations it was supposed to produce have arrived, possibly many turns later --- concluding a sub-goal plan requires everything that happened inside the sub-goal. Second, the environment supplies no per-action reward, and in the lifelong setting may supply none at all. AdMem resolves both at once: the actor must state the expected purpose or result of every action \emph{before} taking it, and the critic buffers the action until its outcomes land and then judges it against that expectation.

Within the actor, the two non-procedural memory types are surfaced in different structures, shown inside the actor box in \Cref{fig:interaction_diagram}. The \textbf{semantic heap} is an unordered, priority-ranked pool: environment facts have no natural sequence, are read by relevance to the current context, and the highest-scoring ones are surfaced first. The \textbf{episodic stack} is ordered and scoped: event summaries are pushed as the task descends into sub-goals and read most-recent-first, so the actor sees the events of the sub-goal it is currently inside before older ones, and a scope's episodic entries are compressed together when that scope is concluded (\Cref{sec:stm}). Both structures draw on one fixed prompt allowance, so what they expose is what the actor gets to reason over.

\subsection{Pipeline and expectation-grounded credit assignment}\label{sec:pipeline}
\Cref{alg:actor,alg:critic} in \Cref{sec:appendix-alg} give the two loops, and \Cref{fig:interaction_diagram} the information flow between them.

The actor owns all communication with the environment and consumes what the store returns; besides acting, it can call planning tools that both structure the task and delimit the units the critic will later judge. The critic owns the procedural write path: for each action it receives the pre-action context, the action, the actor's committed expectation, and the realized outcome, then condenses these, proposes a better alternative, and labels the record with a reward $r_k\in\{0,1\}$ according to whether the outcome met the expectation and advanced the task. Two properties distinguish this from writing memory from task-level verdicts. First, the reward is \emph{per action}: an action taken thirty steps in is judged on its own outcomes, so a good action inside a failed task is not penalized and a lucky one inside a successful task is not rewarded. Second, it requires \emph{no environment signal}, coming instead from a comparison against a commitment made before acting, available in any environment that returns observations. That the standard is fixed before the outcome is known is what makes this more than after-the-fact self-assessment.

The planning tools give the critic well-formed units to wait for (full tool specifications in \Cref{sec:appendix2}). \textbf{Plan for the task} is written before the first action and its entry is emitted at the task's end, once every observation bearing on the plan has arrived; \textbf{add sub-goal} and \textbf{conclude sub-goal} push and pop a scope, and the sub-goal planning action's entry takes everything between push and pop as its observations; \textbf{act} carries an explicit expectation and its entry is emitted after the immediate observation; \textbf{think} emits reasoning into short-term memory and changes nothing. Pushing and popping thus lets the actor explore the environment as a tree while giving the pending queue a natural closing condition: a record is judgeable exactly when its scope closes. Whenever the critic emits a reward it also forwards the identities of the procedural entries that were retrieved for that action --- the only channel by which the store learns whether its own recommendations were any good.

\subsection{Short-term memory: scoping and compaction}\label{sec:stm}
The actor needs to attend to the current goal without re-reading the whole task; the critic needs a delimited execution trace to summarize. Both are served by one structure borrowed from program execution: a running stack. The live frame holds the current sub-goal and the plan the actor made for it; turns outside it are compressed into summaries by the actor's own LLM, so the actor sees the current scope at full resolution and everything above it compressed, and resumes the parent objective when the frame is popped. Sub-goal and task boundaries are also scope boundaries in the memory-safety sense. Popping a sub-goal compresses its turns, so raw observations from a concluded scope are no longer verbatim in the actor's context; ending a task clears short-term memory entirely, and nothing survives into the next task except what the critic or the semantic/episodic summarizer wrote to the long-term store. Cross-task carryover is therefore mediated by summarization rather than by raw trajectory, which bounds context growth and, as \Cref{sec:threat} discusses, what raw environment text can reach a later task.

\subsection{Long-term memory: retrieval, utility, and life cycle}\label{sec:memory-management}
The store holds $M$ and owns both retrieval and the entry life cycle.
\begin{itemize}[itemsep=1pt, topsep=2pt, parsep=0pt, partopsep=0pt]
    \item \textbf{Retrieval.} Semantic and episodic entries are retrieved by dense similarity to the current context. Procedural entries are ranked by a product of their estimated utility and their context similarity, defined below, so that an entry must be both relevant and previously effective in order to be surfaced. Procedural entries are additionally hard-scoped by the action space they were recorded under: an entry is eligible only when the set of legal tools matches, so advice formed in one action space is never surfaced in another.
    \item \textbf{Consolidation.} Two procedural entries are merged into one when the retriever cannot distinguish them: we score a pair by how often they are retrieved \emph{together} among the rounds in which either one is retrieved, and merge above $1-\epsilon_{\mathrm{m}}$, once at least one of the two has accumulated enough retrievals for that rate to mean something. Scoring on co-retrieval rather than on raw round-by-round agreement is what keeps two merely cold entries from looking mutually indistinguishable. Semantic entries are consolidated differently --- an LLM edits the existing entry when a new fact refines or contradicts it, rather than appending a second entry.
    \item \textbf{Forgetting.} Once an entry has been resident long enough for its statistics to be meaningful, it is removed from the retrieval index if its retrieval frequency falls below a threshold $\epsilon_{\mathrm{e}}>0$. Because ranking is utility-weighted, an entry whose utility estimate has fallen stops being retrieved and is then evicted; low utility causes forgetting through the retrieval channel rather than by a direct reward threshold. This is the system's forgetting policy, and its absence is, we argue in \Cref{sec:experiments}, a large part of why unmanaged procedural memory ends up below a memoryless baseline.
    \item \textbf{Provenance.} Each procedural entry retains its originating context and action, the critic's reflection, its binary success label, and its current utility, so it is traceable to the action that produced it and inspectable as text --- which is how the failure modes in \Cref{sec:experiments} were identified.
\end{itemize}

Retrieval is thus the hinge of the life cycle: consolidation and forgetting are both computed from retrieval statistics, so we estimate, rather than assume, how useful each entry is. We give each procedural entry $m\in M_P$ a scalar utility $v_m\in[0,1]$ measuring whether following its suggestion helps. At an actor step with context $c_t$ and retrieved entries $M_t = [m_t^{(1)},\dots,m_t^{(n)}]$ the critic eventually returns $r_t\in\{0,1\}$, and we model that reward as a noisy-OR (\Cref{sec:appendix-em}): the action succeeds if \emph{any} retrieved entry helped, or if it would have succeeded unaided with probability $p_0$. Entry $i$ helps with probability $v_{m_t^{(i)}}\,\psi(c_t)^\top\psi(c^{(i)})$, where $\psi$ is the retriever's normalized embedding map, $c^{(i)}$ is the stored context of entry $m_t^{(i)}$, and the similarity is clipped to $[0,1]$. An entry therefore helps only if its advice is sound \emph{and} its originating situation resembles the present one, which is what that product encodes; the model has one parameter per entry plus the global $p_0$, which is what makes it affordable to keep online for a growing store.

Each observed reward licenses an update to $v_{m_t^{(1)}},\dots,v_{m_t^{(n)}}$, which we perform by ascending the noisy-OR log-likelihood one observation at a time; \Cref{sec:appendix-em} gives the model and the update in full. The update is asymmetric by construction, since a noisy-OR emits $0$ only if every cause failed: a failed action drives all of its retrieved entries down, each by an amount scaled by how well its stored context matched, while a successful one distributes credit in proportion to how likely each entry was to be the cause. Retrieval then selects the top-$n$ entries by
\begin{equation}\label{eq:score}
\mathrm{score}(m \mid c_t) \;=\; v_m\,\psi(c_t)^\top \psi(c_m),
\end{equation}
the same quantity the reward model uses --- the retriever ranks by the model's own estimate of the probability that the entry will help here. Newly written entries start at $v_m=1$, in the spirit of optimism in the face of uncertainty: an untried entry gets a chance to be retrieved so that it can acquire the statistics that will keep or evict it.

\section{Experiments}\label{sec:experiments}

\begin{table}[t]
\centering\resizebox{1.0\linewidth}{!}{
\begin{tabular}{c|cc|cc|cc|cc}
\toprule
Domain  &  \multicolumn{2}{c|}{AlfWorld} &  \multicolumn{2}{c|}{BabyAI} &  \multicolumn{2}{c|}{Jericho} &  \multicolumn{2}{c}{PDDL}  \\ \midrule
Number of tasks &  \multicolumn{2}{c|}{134} &  \multicolumn{2}{c|}{112} &  \multicolumn{2}{c|}{20} &  \multicolumn{2}{c}{60} \\
\midrule
Metric & C & P & C & P & C & P & C & P\\
\midrule
ReAct  & 49.3\%  & 0.6586 & 97.3\% & 0.9732 & 40.0\% & 0.4002 & 33.3\% & 0.5537\\
AWM & 47.0\%  & 0.5467 & 92.9\% & 0.9375 & 10.0\% & 0.2227 &43.3\% & 0.6407\\
\textbf{AdMem}  & \textbf{63.4\%}  & \textbf{0.7755} &\textbf{100.0\%} & \textbf{1.0000} &\textbf{45.0\%}& \textbf{0.5569}&\textbf{76.7\%} & \textbf{0.9069}\\
\midrule
Domain  & \multicolumn{2}{c|}{\shortstack[c]{Tool-query\\Academic}} & \multicolumn{2}{c|}{\shortstack[c]{Tool-query\\Weather}}  & \multicolumn{2}{c|}{WebShop} & \multicolumn{2}{c}{ScienceWorld} \\
\midrule
Number of tasks & \multicolumn{2}{c|}{20} & \multicolumn{2}{c|}{20} & \multicolumn{2}{c|}{251} & \multicolumn{2}{c}{90} \\
\midrule
Metric & C & P & C & P & C & P & C & P\\
\midrule
ReAct  & 70.0\%  & 0.8365 & 60.0\% & 0.8565 & \textbf{42.2\%} & 0.7507 & \textbf{58.9\%} & 0.8258\\
AWM & 55.0\%  & 0.8185 & 25.0\% & 0.7529 & 29.1\% & 0.7012 & 31.1\% & 0.6324\\
\textbf{AdMem}  &
\textbf{95.0\%}  & \textbf{0.9671} & \textbf{75.0\%} & \textbf{0.9036}& 41.4\% & \textbf{0.7645} & 56.7\% & \textbf{0.8447}\\
\bottomrule
\end{tabular}}
\caption{Task completeness (C, the fraction of tasks solved) and average progress (P, the mean of the per-task progress score $p_i\in[0,1]$) on the eight AgentBoard domains, using Claude Haiku 4.5 as the backbone for every method. Bold marks the best value in each column. Each domain is a fixed task set of the size given in the second row, so completion counts are recoverable exactly from C (e.g.\ $63.4\%$ of $134$ is $85$ tasks); we use those counts for the Fisher exact tests reported in the text. Each configuration is streamed once over the task set in a single lifelong episode. On the three 20-task domains a single task is worth $5.0$ points of C, which is why we base per-domain conclusions on the tests rather than on the point differences alone.}
\label{tab:results2}
\end{table}

\begin{table}[t]
\centering\small
\begin{tabular}{lcccc}
\toprule
Method & Tasks solved / 707 & 95\% CI on C & Macro-avg.\ C & Domains $\geq$ ReAct \\
\midrule
ReAct & 388 \quad (54.9\%) & [51.2, 58.5] & 56.4\% & --- \\
AWM & 312 \quad (44.1\%) & [40.5, 47.8] & 41.7\% & 1 of 8 \\
\textbf{AdMem} & \textbf{441} \quad (\textbf{62.4\%}) & \textbf{[58.7, 65.9]} & \textbf{69.2\%} & \textbf{6 of 8} \\
\bottomrule
\end{tabular}
\caption{Aggregate view of \Cref{tab:results2} over all $707$ tasks. Intervals are Wilson $95\%$ binomial intervals on the pooled completion rate. Macro-avg.\ C weights the eight domains equally rather than by task count; we report both because WebShop and ScienceWorld together hold $48\%$ of the tasks. Against ReAct, the pooled difference is significant for AdMem ($+53$ tasks, two-sided Fisher exact $p=5\times10^{-3}$) and significant in the \emph{negative} direction for AWM ($-76$ tasks, $p=7\times10^{-5}$). Because all three methods are run over the same $707$ tasks, both comparisons are paired; both remain significant under the least favorable exact McNemar test consistent with these margins ($p=0.03$ and $p=5\times10^{-3}$ respectively, \Cref{sec:appendix-stats}).}
\label{tab:aggregate}
\end{table}

\begin{table}[t]
\centering\small
\begin{tabular}{c|cc|cc|cc}
\toprule
Domain  & \multicolumn{2}{c|}{\shortstack[c]{Jericho\\First epoch}}  & \multicolumn{2}{c|}{\shortstack[c]{Jericho\\Second epoch}} & \multicolumn{2}{c}{\shortstack[c]{Jericho\\Third epoch}} \\
\midrule
Metric & C & P & C & P & C & P \\
\midrule
ReAct &  40.0\%  & 0.4002&  - & - & - & - \\
STP & 40.0\%& 0.4553&  - & - & - & - \\
LTP  & 20.0\% & 0.4257& 25.0\% & 0.4775 & 30.0\% & 0.4743\\
\textbf{AdMem} &  \textbf{45.0\%}& \textbf{0.5569}& \textbf{50.0\%} & \textbf{0.6578}& \textbf{60.0\%}& \textbf{0.6820}\\ \bottomrule
\end{tabular}
\caption{Component ablation on Jericho (20 tasks), Claude Haiku 4.5. \textbf{STP}: short-term planning and context management only, with all long-term recall removed. \textbf{LTP}: the long-term procedural store only --- per-step retrieval with critic-generated reflections and the reward-based utility retained --- with short-term planning and context management removed. \textbf{AdMem}: all components. STP and ReAct are single-epoch by construction, since neither carries anything across tasks. For LTP and AdMem the task set is streamed again each epoch without resetting the long-term store, so the epoch axis measures accumulation rather than repetition of a fixed store; entry utilities carry over with the store, so an entry that proved unhelpful in an earlier epoch enters the next one already down-weighted.}
\label{tab:results}
\end{table}

\subsection{Setup}
We evaluate AdMem under streaming multi-task conditions intended to approximate a deployed task-solving agent: the agent meets a domain's tasks one after another in a single lifelong episode, and its long-term store is never reset between tasks. We use the eight domains bundled with AgentBoard \citep{ma2024agentboard}, spanning embodied household tasks (AlfWorld), grid-world instruction following (BabyAI), interactive fiction (Jericho), symbolic planning (PDDL), tool calling (Tool-query Academic and Weather), web navigation (WebShop), and scientific experimentation (ScienceWorld). Within a domain every method receives the same system prompt (\Cref{sec:appendix2}) and the unmodified AgentBoard task instructions, and is run over the full task set. We report AgentBoard's two native metrics: task completeness $C$, the fraction of tasks solved, and average progress $P$, the mean over tasks of a per-task progress score $p_i\in[0,1]$ with $p_i=0$ for no progress and $p_i=1$ for completion.

Claude Haiku 4.5 is the backbone for the actor, the critic, and the semantic/episodic summarizer in every configuration, including the baselines, so all comparisons hold the backbone fixed; decoding is greedy (temperature $0$, top-$p$ $1$) and each task is capped at $30$ environment steps, as recorded in the run configuration. Long-term retrieval is dense retrieval over the stored context field of each entry using \texttt{text-embedding-3-small} embeddings, L2-normalized so that the inner product is a cosine. Procedural entries are ranked by the utility-weighted score of \Cref{eq:score} and the top $n=3$ above an inclusion threshold of $0.1$ are returned; episodic entries are ranked by similarity alone, also with $n=3$, while the consolidated semantic entry is injected in full. The life-cycle constants of \Cref{sec:memory-management} are fixed in the implementation rather than set per domain: $\epsilon_{\mathrm{e}}=\epsilon_{\mathrm{m}}=0.01$, with an entry becoming eligible for eviction only after $500$ retrieval rounds of residence and for merging after $100$; the sub-goal stack is capped at depth $3$, and the actor's short-term window holds $30$ messages before the oldest turns of the live frame are compressed. The utility step size is $\eta=10^{-3}$, new entries enter at $\hat v_m=1$, and the implementation runs the $p_0=0$ special case of the reward model (\Cref{sec:appendix-em}).

We compare against two baselines. \textbf{ReAct} \citep{yao2022react} is the no-cross-task-memory control: it interleaves reasoning and acting within a task and starts every task from scratch. \textbf{AWM} \citep{wang2024agentworkflowmemory} is the closest published system to our setting --- procedural memory induced online from the agent's own trajectories and reused on later tasks --- and is therefore the informative comparison for the question this paper asks. We do not port conversational-memory systems such as Mem0 \citep{chhikara2025mem0buildingproductionreadyai}, MemGPT \citep{packer2024memgptllmsoperatingsystems}, or MIRIX \citep{wang2025mirixmultiagentmemoryllmbased}: those retrieve over accumulated dialogue and documents, and adapting them here would require choosing what a memory unit is for an environment step and what the retrieval query is at action time, so the numbers would characterize our port rather than the original systems. The role a reflection-memory baseline would play is instead filled from inside our own system by the \textbf{LTP} ablation of \Cref{tab:results}, which retrieves critic-written reflections at every step with no planning or scoping structure around them.

\subsection{Cross-task memory is not free}
The headline of \Cref{tab:results2} is not only that AdMem leads, but where the cross-task memory baseline ends up. AWM solves fewer tasks than the memoryless ReAct control in seven of the eight domains, and $312$ of $707$ tasks against ReAct's $388$ --- a $10.7$-point drop in pooled completeness, significant under a two-sided Fisher exact test ($p=7\times10^{-5}$, \Cref{tab:aggregate}). The same pattern appears inside our own system: the LTP ablation of \Cref{tab:results}, which keeps the procedural store, critic reflections, and reward-based utility but removes planning and scoping, solves $4$ of $20$ Jericho tasks against ReAct's $8$.

AWM is not mis-ported: it beats ReAct on PDDL in this very table ($43.3\%$ versus $33.3\%$), so the mechanism does fire where the domain supplies reusable structure \citep{zheng2024synapse,wang2024voyager}. Two properties of induction-from-successes explain the rest. First, AWM induces workflows online from its own trajectories, so a wrong induction --- a step order that happened to work once, a precondition read as universal --- is written with the same confidence as a right one, and a single pass gives no held-out signal that would catch it, a gap self-judgement cannot close \citep{huang2024selfcorrect,stechly2025selfverification}; AdMem's critic shares that limit, but its entries remain revisable as later outcomes update their utility. Second, AWM has no removal policy: a badly generalizing workflow stays and is re-retrieved for the rest of the run, so one bad induction is a persistent rather than transient cost. Under a coarse success signal such inaccuracies compound, and regulating entry quality rather than accumulating more of it recovers the loss \citep{xiong2026memory}; an inapplicable entry distracts rather than sitting inert \citep{shi2023distracted}. AWM was validated on web navigation, where consecutive tasks share page structure; most AgentBoard domains lack that regularity. A streaming benchmark of self-evolving memory reports the same pattern for this class of method: AWM falls below a ReAct control on its multi-turn suite for three of four backbones ($0.49$ against $0.57$ mean success on Claude 3.7 Sonnet) and below the no-memory baseline on its single-turn suite for all four \citep{evomemory2025}. The design question is therefore not how to store more experience but which to keep --- which motivates the per-action credit assignment and entry life cycle of \Cref{sec:memory-management}.

\subsection{Main results}
AdMem solves $441$ of $707$ tasks against $388$ for ReAct and $312$ for AWM (\Cref{tab:aggregate}), and matches or exceeds ReAct on completeness in six of eight domains while achieving the best average progress in all eight. We treat the pooled comparison over all $707$ tasks as the primary endpoint and the eight per-domain tests as a multiplicity-corrected breakdown. Because each domain is a fixed-size task set, completion counts are exact and we test each with a two-sided Fisher exact test. Pooled, the gain is $+7.5$ points ($p=5\times10^{-3}$), with a Wilson $95\%$ interval of $[58.7, 65.9]$ against $[51.2, 58.5]$ for ReAct. Since every method is run over the same task set that comparison is paired, and over \emph{every} pairing consistent with the observed margins the least favorable exact McNemar test still gives $p=0.03$ (\Cref{sec:appendix-stats}), so the pooled conclusion does not rest on the pairing being favorable. Across the eight domain tests, PDDL survives Holm--Bonferroni correction at $\alpha=0.05$ --- $46$ of $60$ tasks against $20$ of $60$, $+43.3$ points, $p=3\times10^{-6}$ against a threshold of $6.3\times10^{-3}$, and $p=5\times10^{-4}$ under the same worst-case pairing --- while AlfWorld, $85$ of $134$ against $66$ of $134$ ($+14.2$ points, $p=0.03$), is nominally significant but does not survive correction. We therefore rest the domain-level claim on PDDL. Both have transferable structure: PDDL tasks share an action schema and AlfWorld tasks reuse the same household affordances, so a reflection about a mis-ordered pick-and-place transfers to a later task.

On WebShop and ScienceWorld, AdMem does not lead on completeness. The margin in each case is two tasks --- $104$ versus $106$ of $251$ and $51$ versus $53$ of $90$ --- indistinguishable from zero ($p=0.93$ and $p=0.88$), with Wilson intervals that overlap almost completely ($[35.5, 47.6]$ versus $[36.3, 48.4]$ on WebShop), while AdMem attains the higher average progress in both ($0.7645$ versus $0.7507$; $0.8447$ versus $0.8258$). We read these as parity on completion with a small edge on partial progress, and note the contrast with AWM, which loses $13.1$ and $27.8$ points on the same task sets: what these columns support is that AdMem's structure prevents the collapse unmanaged cross-task memory produces where transfer is weak, not that it gains there. These are also the domains where transferable structure is thinnest --- each WebShop task targets a different product with a different attribute set, so carryover is closer to distraction than to a reusable recipe, the regime where top-ranked but inapplicable retrievals measurably hurt \citep{cuconasu2024powerofnoise}.

The three 20-task domains behave as expected but carry little weight on their own --- the $+25.0$ points on Tool-query Academic ($19$ of $20$ against $14$) reaches only $p=0.09$. \Cref{tab:aggregate} gives both the pooled and the equally-weighted domain average rather than the flattering one: they differ ($+7.5$ against $+12.8$ points) because WebShop and ScienceWorld hold $48\%$ of the tasks, and the pooled figure is the conservative reading.

\subsection{Which components matter}
\Cref{tab:results} decomposes the system on Jericho into (i) acting and planning with short-term memory management (STP) and (ii) the long-term procedural store (LTP), and additionally streams the task set for three epochs to test whether accumulation pays off over a longer horizon. The two halves behave asymmetrically. STP raises average progress over ReAct ($0.4553$ against $0.4002$) at equal completeness, and is single-epoch by construction: nothing crosses task boundaries, so there is no horizon over which it could improve. LTP is the weaker half in isolation --- $4$, $5$, then $6$ of $20$ tasks across the three epochs, below ReAct's $8$ at every epoch. It improves with accumulation, but the trend does not carry it to the memoryless baseline within three epochs, so the long-term store on its own is net-negative on this domain rather than a transient warm-up cost.

Assembled, the two halves reverse that: AdMem improves monotonically, $9\to10\to12$ of $20$ tasks, ending $+20.0$ points above ReAct. This is the paper's clearest claim about the architecture: the procedural store is only useful inside the planning and scoping structure, which turns a net-negative store into a net-positive one. It also answers the simpler-baseline question: LTP retains the critic's reflections \emph{and} the reward-derived utility, so it is if anything a strengthened reflection-memory agent, and it is the weakest arm in the table. With $20$ tasks the per-epoch differences are individually underpowered (LTP versus AdMem at epoch three is $6$ against $12$, $p=0.11$), and the contrast is between two halves of the system rather than individual mechanisms.

\subsection{Cost}
The design prioritizes long-horizon performance over minimal per-step cost. Multi-threading keeps memory encoding, management, and the critic's write path off the actor's critical path, but the pipeline still spends extra planning turns and background LLM calls. As a rough worst case, per-step wall-clock is about $2\times$ the vanilla LLM agent's with about $3\times$ the LLM calls --- a bound set by the easiest domains, where any fixed overhead looks large in ratio. Extra memory-side calls are not the differentiator: AWM and the LTP ablation also pay them, and both land below ReAct.

\section{Memory as an attack surface}\label{sec:threat}
Keeping memory in readable text is what makes the diagnosis in \Cref{sec:experiments} possible, and the same property makes the store a durable, attacker-reachable channel. The dataflow follows from the architecture: environment text enters the critic prompt as an action's realized outcome, becomes a procedural entry, and is later retrieved into the actor's context before it acts --- so any text the environment can emit reaches a later task's action selection, and unlike a payload confined to one context window a written entry persists and is re-retrieved. This is indirect prompt injection \citep{greshake2023youvesignedforcompromising} applied to a store the agent writes itself, close to demonstrated poisoning of agent memory and RAG stores \citep{chen2024agentpoison}. Two of our own choices bear on this: $v_m=1$ grants a new entry maximum retrieval priority before any evidence exists, and the utility signal is produced by an LLM reading environment text. The life cycle also supplies the defenses --- unhelpful entries are evicted, provenance is retained, tool-set scoping bounds reach, and a readable store can be audited --- so eviction is a defense and $v_m=1$ a vulnerability within one policy. \Cref{sec:appendix-threat} gives the full dataflow, defenses, and caveats; we report no adversarial evaluation and make no robustness claim.

\section{Limitations}
Our evidence is one run per configuration on one backbone. We compensate where the data allows --- completion counts on fixed task sets support Fisher exact tests and Wilson intervals --- but this is not a full seed sweep, and the three 20-task domains are underpowered by construction. A caveat specific to the memory arms is that each is one coupled stream, so their binomial intervals are anti-conservative; we therefore report the variance inflation each load-bearing conclusion tolerates before crossing $\alpha=0.05$, a design effect of $4.2$ for the negative result and $5.9$ for the PDDL gain (\Cref{sec:appendix-stats}). The component ablation separates the acting/planning half from the long-term half on one domain, so we do not claim how much of the gain is attributable to expectation-grounded reward, to utility-based eviction, or to context scoping.

\section{Conclusion}
We asked which experience a task-solving agent should keep. Procedural memory induced from successes leaves an agent below a memoryless control on seven of eight AgentBoard domains; AdMem answers with per-action credit against a pre-committed expectation and a per-entry utility governing the entry life cycle. The mechanism that lets memory help where transfer exists is the one that keeps it from hurting where it does not.

\phantomsection\label{endofmainbody}
\bibliography{references}
\bibliographystyle{plainnat}

\appendix
\crefalias{section}{appendix}

\section{The actor and critic loops}\label{sec:appendix-alg}

\begin{algorithm}[H]
\caption{Actor loop}
\label{alg:actor}
\begin{algorithmic}
\State Initial short-term memory $s_0$. Round number $t\leftarrow 0$.
\While{active}
    \State $t\leftarrow t+1$. Receive observation $o_t$ from the user / environment.
    \State Send $o_t$ to the critic.
    \State Send the context $c_t = (s_{t-1}, o_t)$ to the memory store to write semantic and episodic entries.
    \State Query the store for procedural entries $M_t = [m_t^{(1)},\dots,m_t^{(n)}]$ and for semantic and episodic entries $M'_t$, given $c_t$.
    \State Choose action $a_t$ from $(c_t, M_t, M'_t)$ and \emph{commit to} an expected outcome $\hat{o}_{t+1}$ for it. Update short-term memory, $s_t \leftarrow p(s_{t-1}, o_t, a_t)$.
    \State Send the pending record $k = (c_t, a_t,\hat{o}_{t+1}, M_t)$ to the critic.
\EndWhile
\end{algorithmic}
\end{algorithm}

\begin{algorithm}[H]
\caption{Critic loop (runs concurrently with \Cref{alg:actor})}
\label{alg:critic}
\begin{algorithmic}
\State Pending queue $Q\leftarrow []$.
\While {active}
    \If{a pending record $k$ arrives from the actor}
        \State put $k$ into $Q$.
    \EndIf
    \If{an observation $o$ arrives from the actor}
        \For {every pending record $k=(c_k,a_k,\hat{o}_{k},M_k)$ in $Q$}
            \If {$o$ is relevant to $k$}
                \State Append $o$ to $k$.
                \If {$k$ has received all the observations needed to judge it}
                    \State Compare the realized outcome against the committed expectation $\hat o_k$;
                    \State emit reward $r_k\in\{0,1\}$ and reflection $f_k$.
                    \State Write the procedural entry $(c_k,a_k,\hat{o}_{k},r_k,f_k)$ to the store, with $v\leftarrow 1$.
                    \State Send $(M_k,r_k)$ to the store, crediting the entries that were retrieved
                    \State for this action, to update their utilities (\Cref{sec:memory-management}).
                    \State Remove $k$ from $Q$.
                \EndIf
            \EndIf
        \EndFor
    \EndIf
\EndWhile
\end{algorithmic}
\end{algorithm}

\paragraph{What ``relevant'' means in \Cref{alg:critic}.} An observation $o$ is relevant to a pending record $k$ exactly when $o$ arrives inside the scope that $k$ opened. For an \textbf{act} record that is the immediately following observation; for an \textbf{add sub-goal} record it is every observation between that push and the matching pop; for a \textbf{plan for the task} record it is every observation in the task. Relevance is therefore decided by the sub-goal stack rather than by a model call, and $k$ becomes judgeable exactly when its scope closes, which is what bounds the pending queue. Note also that the semantic and episodic write in \Cref{alg:actor} is issued before the retrieval query at the same step, so a summary of the current observation can be returned by that query; this is redundant rather than circular, since the actor has already seen the observation being summarized, and it does not affect procedural retrieval, which is fed only by the critic's asynchronous write path.

\section{Utility updates for the procedural entries}
\label{sec:appendix-em}
We write out the reward model and the update referred to in \Cref{sec:memory-management}. Fix an actor step with context $c_t$, retrieved procedural entries $M_t=[m_t^{(1)},\dots,m_t^{(n)}]$, and observed reward $r_t\in\{0,1\}$. The noisy-OR model is
\begin{equation}\label{eq:noisyor}
r_t \mid c_t \;=\; l_t^{(1)} \vee \cdots \vee l_t^{(n)} \vee l_t^{(0)},
\qquad
l_t^{(i)} \sim \mathcal{B}\!\left(v_{m_t^{(i)}}\,\psi(c_t)^\top \psi(c^{(i)})\right),
\qquad
l_t^{(0)} \sim \mathcal{B}(p_0),
\end{equation}
where $\mathcal{B}(p)$ is Bernoulli with mean $p$, the latent $l_t^{(i)}\in\{0,1\}$ indicates that entry $i$ helped, and $l_t^{(0)}$ that the action would have succeeded with no memory at all. The single global parameter $p_0$ for the memory-free success rate is estimated alongside the $v_m$. Abbreviate the per-entry help probability under the current estimates as
\[
q_i \;=\; \hat v_{m_t^{(i)}}\,\sigma_i, \qquad \sigma_i \;=\; \big[\psi(c_t)^\top\psi(c^{(i)})\big]_{[0,1]},
\qquad q_0 \;=\; \hat p_0 ,
\]
where $[\cdot]_{[0,1]}$ clips to the unit interval. Under the noisy-OR of \Cref{eq:noisyor} the reward is $1$ unless every cause fails, so
\begin{equation}\label{eq:likelihood}
\Pr[r_t = 1 \mid c_t, M_t] \;=\; 1 - \prod_{i=0}^{n}(1-q_i).
\end{equation}

\paragraph{E-step.} The latent variables are the indicators $l_t^{(i)}$. Their posteriors given the observed reward are
\begin{equation}\label{eq:estep}
\gamma_i \;=\; \Pr\!\big[l_t^{(i)}=1 \mid r_t\big] \;=\;
\begin{cases}
\dfrac{q_i}{\,1 - \prod_{j=0}^{n}(1-q_j)\,}, & r_t = 1,\\[2ex]
0, & r_t = 0 .
\end{cases}
\end{equation}
The $r_t=0$ case is exact rather than an approximation: a noisy-OR outputs $0$ only if every cause failed, so observing $r_t=0$ deterministically sets every $l_t^{(i)}=0$. This is the asymmetry that gives the estimator its behavior --- a failed action drives all of its retrieved entries down together, while a successful one distributes credit in proportion to how likely each entry was to be the cause. An entry retrieved alongside a strong entry receives little credit for a success, and full blame for a failure.

\paragraph{M-step.} Each retrieved entry's utility is moved towards its responsibility discounted by the similarity that gated it, since $\gamma_i$ estimates $\Pr[l^{(i)}=1]=v_{m}\sigma_i$ and we want $v_m$:
\begin{equation}\label{eq:mstep}
\hat v_{m_t^{(i)}} \;\leftarrow\; \Big[(1-\eta)\,\hat v_{m_t^{(i)}} \;+\; \eta\,\frac{\gamma_i}{\sigma_i}\Big]_{[0,1]},
\qquad
\hat p_0 \;\leftarrow\; \big[(1-\eta)\,\hat p_0 + \eta\,\gamma_0\big]_{[0,1]},
\end{equation}
with step size $\eta\in(0,1]$. Entries not retrieved at step $t$ are untouched, so the cost of an update is $O(n)$ in the number of retrieved entries and independent of the size of the store. Newly written entries enter with $\hat v_m = 1$.

\paragraph{Gradient form.} \Cref{eq:mstep} and a stochastic-gradient step on the same objective are interchangeable in this system, and our implementation uses the latter:
\begin{equation}\label{eq:gradform}
\hat v_{m_t^{(i)}} \;\leftarrow\; \Big[\hat v_{m_t^{(i)}} \;+\; \eta\,\frac{\partial}{\partial \hat v_{m_t^{(i)}}}\log \Pr[r_t \mid c_t, M_t]\Big]_{[0,1]},
\end{equation}
which for $r_t=1$ evaluates to $\hat v_i \!+\! \eta\,\sigma_i P_0/\big((1-q_i)(1-P_0)\big)$ with $P_0=\prod_{j}(1-q_j)$, and for $r_t=0$ to $\hat v_i \!-\! \eta\,\sigma_i/(1-q_i)$. Both \Cref{eq:mstep} and \Cref{eq:gradform} ascend \Cref{eq:likelihood} and inherit the same asymmetry; \Cref{eq:gradform} avoids the division by $\sigma_i$ and is the more numerically forgiving of the two. To keep the noisy-OR factors away from the degenerate ends of the unit interval, $q_i$ is clipped to $[0.05, 0.95]$ before use.

\paragraph{The $p_0$ term in practice.} Our implementation fixes $p_0=0$ rather than estimating it, so \Cref{eq:likelihood} reduces to a product over the $n$ retrieved entries. This is exact for the $r_t=0$ branch, whose update $-\eta\,\sigma_i/(1-q_i)$ contains no $p_0$, but over-credits on $r_t=1$: with the unaided route removed, a success must be explained by the retrieved entries, so the credit they receive is an upper bound on what a model with $p_0>0$ would assign. The two settings also agree at initialization, since a new entry enters at $\hat v_m = 1$, the upper clip bound. We keep $p_0$ in the model because it is what makes the noisy-OR the right generative story for a reward that does not require memory to be present, and record the simplification here rather than leaving it implicit.

\paragraph{What the model can and cannot express.} Six properties should be stated plainly. First, this is a \emph{streaming} update: one step per observed reward rather than iterating to convergence on a retained dataset, so the monotone-likelihood guarantee of batch EM does not transfer and we do not claim it. The estimator should be read as tracking a non-stationary quantity --- an entry's usefulness genuinely changes as the rest of the store changes around it --- rather than as maximizing a fixed likelihood. Second, retrieval only surfaces entries whose $\mathrm{score}$ exceeds the inclusion threshold, which bounds $\sigma_i$ away from zero for exactly the entries that get updated. Third, the similarity $\sigma_i$ is supplied by a frozen embedding model and is never itself learned, so the model can only reallocate credit among retrieved entries and cannot discover that its notion of context similarity is wrong.

Fourth, $v_m$ is bounded below at $0$, so the model represents an entry that never helps but not one that actively hurts. This is deliberate rather than an oversight: harm leaves the store through the retrieval channel rather than through a negative score. An entry that keeps co-occurring with failures is driven towards $0$, its \Cref{eq:score} value falls with it, it stops clearing the inclusion threshold, and the frequency-based eviction of \Cref{sec:memory-management} then deletes it --- so a harmful entry and a merely useless one receive the same treatment, which is removal, and a signed utility would only be needed if the system had to rank harmful entries against one another, which it never does. It is also worth being precise about how a failure is charged: by \Cref{eq:gradform} the decrement is $\eta\,\sigma_i/(1-q_i)$, which is \emph{not} uniform over the retrieved set --- an entry whose stored context closely matches the present one loses more than a marginally similar one, and an entry currently estimated to help loses more than one already near zero.

Fifth, the update conditions on retrieval, not on use: an entry the actor ignored is credited or blamed as if it had been followed, a failure with an external cause, such as a broken tool, lowers every entry retrieved for that action, and the noisy-OR treats retrieved entries as independent causes, so overlapping or conflicting entries are not modelled as such. Finally, we read $v_m$ as the model's estimate of the probability that an entry helps here, which is the quantity retrieval ranks by; the causal phrasing is licensed by the generative model rather than measured against an intervention, and we make no calibration claim.

\section{Pairing and clustering robustness of the pooled tests}
\label{sec:appendix-stats}

Two properties of the design are not captured by a Fisher exact test on a $2\times2$ table, and we bound the effect of both rather than assume them away.

\paragraph{Pairing.} All three methods are run over the same $707$ tasks, so the pooled comparison is paired and the appropriate test is McNemar's on the discordant pairs. We do not log per-task outcomes for the baselines, so the pairing itself is unavailable; instead we take the worst case over every pairing consistent with the observed margins. Given the two margins --- $441$ solved for AdMem against $388$ for ReAct --- write $b$ for the tasks AdMem solves and ReAct fails and $c$ for the converse, so that $b-c=53$ is fixed while $b+c$ is not. The exact two-sided McNemar test, a binomial test on $b$ out of $b+c$ at rate $\tfrac12$, is least significant when the discordant count $b+c$ is largest, so we maximize it subject to $b\le\min(441,\,707-388)$ and $c=b-53\ge0$. The extreme admissible pairing is $b=319$, $c=266$, giving $p=0.031$. The same construction for AWM against ReAct, with $c-b=76$, has extreme point $b=312$, $c=388$ and gives $p=4.6\times10^{-3}$; for PDDL, $b=40$, $c=14$ and $p=5.4\times10^{-4}$. Every pairing consistent with the reported margins is therefore at least this significant, so neither pooled conclusion nor the PDDL result depends on the unobserved pairing being favorable. Applying the same construction to AlfWorld gives $p=0.096$, which is one more reason we do not rest a claim on that domain.

\paragraph{Clustering within a stream.} Tasks in a memory arm are not independent trials: they share one evolving store, so a single badly generalizing entry can correlate failures across the tasks that follow it. This inflates the true variance of a pooled rate by a design effect $D$, equivalently reducing the effective sample size to $n/D$, and it is precisely the mechanism \Cref{sec:experiments} argues is operating. Rather than assume $D=1$, we report the breakdown point: dividing the observed two-proportion $z$ by $\sqrt{D}$, the negative result for workflow-induction memory ($z=-4.04$) crosses $\alpha=0.05$ only at $D=4.2$, and the PDDL gain ($z=4.77$) only at $D=5.9$. A design effect above $4$ within a single stream of a few hundred tasks would require most of the outcome variance to be attributable to stream position, which the per-domain spread in \Cref{tab:results2} does not support. We note that this is a sensitivity analysis and not a substitute for replicated stream orders, which remains the right experiment.

\section{The memory store as an attack surface: full dataflow}
\label{sec:appendix-threat}
This expands \Cref{sec:threat}. The write path is: environment text $\rightarrow$ critic prompt (as the realized outcome of a pending action) $\rightarrow$ procedural entry $\rightarrow$ retrieval index $\rightarrow$ actor context at some later step, possibly in a later task, where the actor is instructed to reason over the retrieved entry before choosing an action. Three properties of that path matter. It is \emph{persistent}: a payload that succeeds once is stored and re-retrieved, so the cost of a single successful injection is amortized over every later retrieval, in contrast to a single-context payload. It is \emph{self-authored}: no external document has to be trusted, because the agent writes the entry itself from text the environment supplied. And it is \emph{reward-coupled}: the utility $v_m$ that governs retrieval, consolidation, and eviction is produced by an LLM critic reading environment text, so an adversary who can shape observations can in principle shape the signal that decides what the store keeps.

Initializing $v_m=1$ is the choice with the sharpest two-sidedness. It is the exploration policy that lets an untried entry acquire the statistics that will keep or evict it, and it is exactly the property an attacker who can cause one entry to be written would want, since maximum retrieval priority is granted before any evidence exists. A pessimistic initialization would blunt that at the cost of never surfacing genuinely useful new entries; we did not evaluate the trade-off and note it as the obvious knob.

The defenses available without adding any new machinery are the life cycle itself. Utility-weighted ranking means an entry that does not help stops being retrieved, and the frequency-based eviction of \Cref{sec:memory-management} then removes it, so a payload that does not produce apparent success decays. Provenance makes every entry traceable to the action, context, and reflection that produced it. Tool-set scoping bounds which later contexts an entry can reach at all: an entry recorded under one action space is ineligible in another, which limits lateral movement across domains. And because the store is human-readable text, an operator can audit and delete, which is not true of a store held in parameters or opaque embeddings.

On personal data specifically, the critic prompt (\Cref{sec:appendix2}) instructs the critic to replace identifying values with placeholders before writing, which limits what crosses a task boundary. This is an instruction-level measure and not a guarantee: it depends on the critic complying, it does not cover what a summarizer writes into semantic or episodic entries, and it does not address the contextual-integrity question of which information is appropriate to carry into which later context \citep{mireshghallah2024confaide}. A deployment with real users would need a mechanism rather than an instruction here. We report no adversarial evaluation and make no robustness claim; the purpose of this section is to state the dataflow and the two design choices that bear on it, so that they are visible rather than found.

\newpage
% The prompt boxes below are single unbreakable \fbox{\parbox} units; without
% \raggedbottom LaTeX stretches the inter-paragraph glue to fill the page around them.
\raggedbottom
\section{LLM Prompts}
\label{sec:appendix2}
Here we include the prompts used by the actor and by the critic. Bold segments are filled with the corresponding information at run time.\\

\fbox{
  \parbox{0.95\linewidth}{\textbf{Actor:}\\\noindent\rule{\linewidth}{0.4pt}\\
\small You are a helpful agent that solves tasks in a given environment. You take turns to interact with the environment and learn about the environment from the observations, until you achieve the goal defined in the task. A log for your previous turns in the task is given below, where planner messages (if exists) marks the plan you made earlier for solving the task and agent\_summary messages provide summarization of one or more turns in the process. \\\\In the next turn you MUST use one of the tools in the \textless tools\textgreater tag. Do not output plain text. Some tools are provided for you to plan for next steps in the task and some tools are provided to interact with the environment. You need to decide which tool to use, but to solve hard task efficiently, careful planning is important. Always make sure you generate valid JSON for using the tool. \\\\Before action, the system will also generate some messages by recalling the system memory. These messages include information from your past interactions with the environment, as well as reflections on your past actions under similar circumstances (for other tasks in the same environment). Please examine these messages and think about their implications for the possible outcomes or your action, before deciding your action.\\\\\textless tools\textgreater\\\hspace*{2em}\textbf{[Memory Tool Descriptions] }\\\textless/tools\textgreater\\\\Below is the description of the task and the environment.\\\\
\hspace*{2em}\textbf{[Task Instructions]}\\
\\Below is the task execution log:\\\\\hspace*{2em}\textbf{[Short Term Memory Output]}\\\\
Below is some memory recall for past experience under similar circumstances:
\\\\\hspace*{2em}\textbf{[Long Term Memory Output]}\\}}

\bigskip
\fbox{
  \parbox{0.95\linewidth}{\textbf{Critic:}\\\noindent\rule{\linewidth}{0.4pt}\\
\small You are a helpful assistant that generates a procedural memory entry from an event. The event is an action (or a plan) taken by an agent when solving a task. The purpose of the procedural memory entry is to provide a description of the context of the action (plan), the action (plan) taken by the agent, the expected outcome of the action (plan), a summary of the actual outcome after the action (plan), and some evaluations and reflections aiming to optimize the action based on the result. The procedural memory should not be too long, but should be helpful in improving the agent's behavior when provided to the agent in a similar situation in the future. Hide private information like user name, account number etc. to protect privacy, for example use [user name] in the place of actual user name. \\\\The context of the action (or plan):\textbf{[context]}\\\\The action (or plan) taken by the agent:\textbf{[action]}\\\\The expected outcome/purpose:\textbf{[expectation]}\\\\The actual outcome:\textbf{[observation]}\\\\
Below is some domain information that helps you understand the task setting.\\\\\hspace*{2em}\textbf{[Domain Information]}\\\\
Now it's your turn to generate the procedural memory entry. To generate, call the tool 'generate\_procedural\_memory' and fill in the arguments with corresponding information.
\\\\\textless tool\_description\textgreater \\
\textit{Name:} generate\_procedural\_memory\\
\textit{Description:} Generate procedural memory entry.\\
\textit{Args:}\\
\hspace*{1em}\textit{context (str):} the action context, a brief and high-level background description summarizing the context where the agent action occurs. \\
\hspace*{1em}\textit{result (str):} a summary of the result of the action or plan, including 1. the agent's expected outcome / achievement; 2. the actual outcome; 3. whether they are consistent.\\
\hspace*{1em}\textit{success (bool):} True or False that the action was successful, considering 1. whether the outcome of the action meets the expectation, 2. whether the outcome offers progress towards the task solving and plan execution.\\
\hspace*{1em}\textit{reflection (str):} Think about what can be learned from the result of the action, specifically what to change in the agent's expectation, and what should the agent do under the same circumstances to better carry out the plan and solve the task. Specifically, think about whether the agent should set up a subgoal first instead of directly interacting with the environment, to better organize the task solving process.\\ \textless /tool\_description\textgreater}\\}\\

We also provide the descriptions for the memory tools we built. Notice that not all memory tools are activated simultaneously to the agent: at the start of a task, the agent can only use plan\_for\_the\_task. Later it can use add\_subgoal if the stack depth of sub-goals is within limit; it can use conclude\_subgoal if there is an active sub-goal; it can use think if the previous turn is an action in the environment, and it can use take\_action always after task planning.\\

\fbox{
  \parbox{0.95\linewidth}{\textbf{Memory Tools:}\\\noindent\rule{\linewidth}{0.4pt}\\
  \small \textit{Name:} plan\_for\_the\_task\\
  \textit{Description:} Plan for a solution. Set the goal and highlight the key points in solving the task, and plan for the task by writing a list of step-by-step instructions. Each instruction can either be high-level subgoal (e.g. learn something about the environment or achieve certain status), or some actions to interact with the environment (e.g. use certain tools in the environment or communicate with the environment). If the task is hard, decompose the task into multiple subgoals that are needed to be attained in line to solve the task.\\
\textit{Args:}\\
\hspace*{1em}\textit{goal (str):}The goal and key points of the current task.\\
\hspace*{1em}\textit{plan (str):} the step-by-step plan for solving the task.\\\noindent\rule{\linewidth}{0.4pt}

\textit{Name:} add\_subgoal\\
\textit{Description:} Add a sub-goal to the current task and make plans to achieve the sub-goal. The sub-goal should admit an observation-based success criterion and should be helpful to solving the final task. Use this method when the task is hard to complete or when you believe the current task plan is not effective to solve the task any more.\\
\textit{Args:}\\
\hspace*{1em}\textit{subgoal (str):} The current sub-goal to be set.\\
\hspace*{1em}\textit{plan (str):} A list of step-by-step instructions that can be followed to achieve the sub-goal.\\\noindent\rule{\linewidth}{0.4pt}
\textit{Name:} conclude\_subgoal\\
\textit{Description:} Conclude the current sub-goal and judge whether it is successfully achieved. You should call the method if the current sub-goal is already achieved or it cannot be achieved anymore. \\
\textit{Args:}\\
\hspace*{1em}\textit{evaluation (str):} A short natural language justification of whether the sub-goal is successful\\
\hspace*{1em}\textit{success (bool):} Whether the sub-goal is successfully achieved.
\\\noindent\rule{\linewidth}{0.4pt}
\textit{Name:} take\_action\\
\textit{Description:} Take an action in the environment. In this method, you can take an action in the environment by passing the action information in the action argument. Write what you expect the action to achieve (or the purpose of the action) in the expectation argument.\\
\textit{Args:}\\
\hspace*{1em}\textit{action (str):} The action to be taken.\\
\hspace*{1em}\textit{expectation (str):} The expected outcome / purpose of the action.
\\\noindent\rule{\linewidth}{0.4pt}
\textit{Name:} think\\
\textit{Description:} Generate some thoughts. This method will not change the environment.\\
\textit{Args:}\\
\hspace*{1em}\textit{thoughts (str):} Some thoughts.\\}}

\end{document}